\documentclass[10pt,twocolumn,letterpaper]{article}
\usepackage{cvpr}      
\usepackage[table,xcdraw,dvipsnames]{xcolor}

\definecolor{cvprblue}{rgb}{0.21,0.49,0.74}
\usepackage[pagebackref,breaklinks,colorlinks,citecolor=cvprblue]{hyperref}
\usepackage[normalem]{ulem}
\usepackage{graphicx}
\useunder{\uline}{\ul}{}
\def\paperID{8596} 
\def\confName{CVPR}
\def\confYear{2024}

\title{Structural Teacher-Student Normality Learning for Multi-Class Anomaly Detection and Localization
}

\author{Hanqiu Deng
and 
Xingyu Li \\
\{hanqiu1, xingyu\}@ualberta.ca\\
University of Alberta\\
}

\begin{document}
\maketitle

\begin{abstract}

Visual anomaly detection is a challenging open-set task aimed at identifying unknown anomalous patterns while modeling normal data. The knowledge distillation paradigm has shown remarkable performance in one-class anomaly detection by leveraging teacher-student network feature comparisons. However, extending this paradigm to multi-class anomaly detection introduces novel scalability challenges. In this study, we address the significant performance degradation observed in previous teacher-student models when applied to multi-class anomaly detection, which we identify as resulting from cross-class interference. To tackle this issue, we introduce a novel approach known as \underline{\textbf{S}}tructural Teacher-Student \underline{\textbf{N}}ormality \underline{\textbf{L}}earning (SNL): (1) We propose spatial-channel distillation and intra-\&inter-affinity distillation techniques to measure structural distance between the teacher and student networks. (2) We introduce a central residual aggregation module (CRAM) to encapsulate the normal representation space of the student network. We evaluate our proposed approach on two anomaly detection datasets, MVTecAD and VisA. Our method surpasses the state-of-the-art distillation-based algorithms by a significant margin of 3.9\% and 1.5\% on MVTecAD and 1.2\% and 2.5\% on VisA in the multi-class anomaly detection and localization tasks, respectively. Furthermore, our algorithm outperforms the current state-of-the-art unified models on both MVTecAD and VisA.

\end{abstract}

\begin{figure}[ht]
  \centering
   \includegraphics[width=\linewidth]{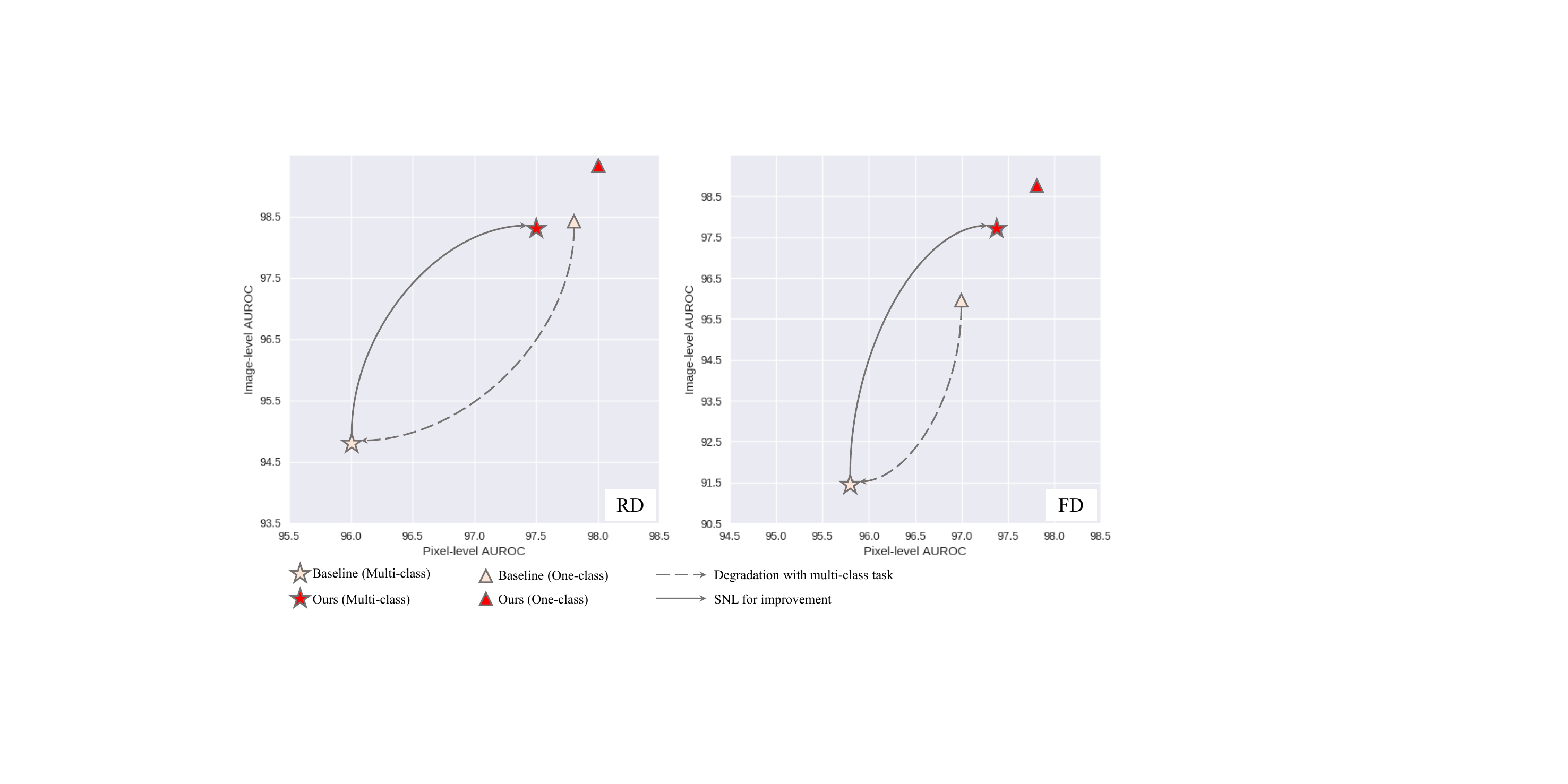}
   \caption{We visualize the performance degradation of one-class teacher-student networks, RD \cite{rd} (left) and FD \cite{fd} (right), in the multi-class anomaly detection task on MVTecAD. Our structural normality learning (SNL) strategy on the teacher-student model shows significant improvement of multi-class anomaly detection and localization on both methods. Besides, SNL can also boost the performance on one-class cases.}
   \label{teaser_mvtec}
\end{figure}

\begin{figure*}[ht]
  \centering
   \includegraphics[width=\textwidth]{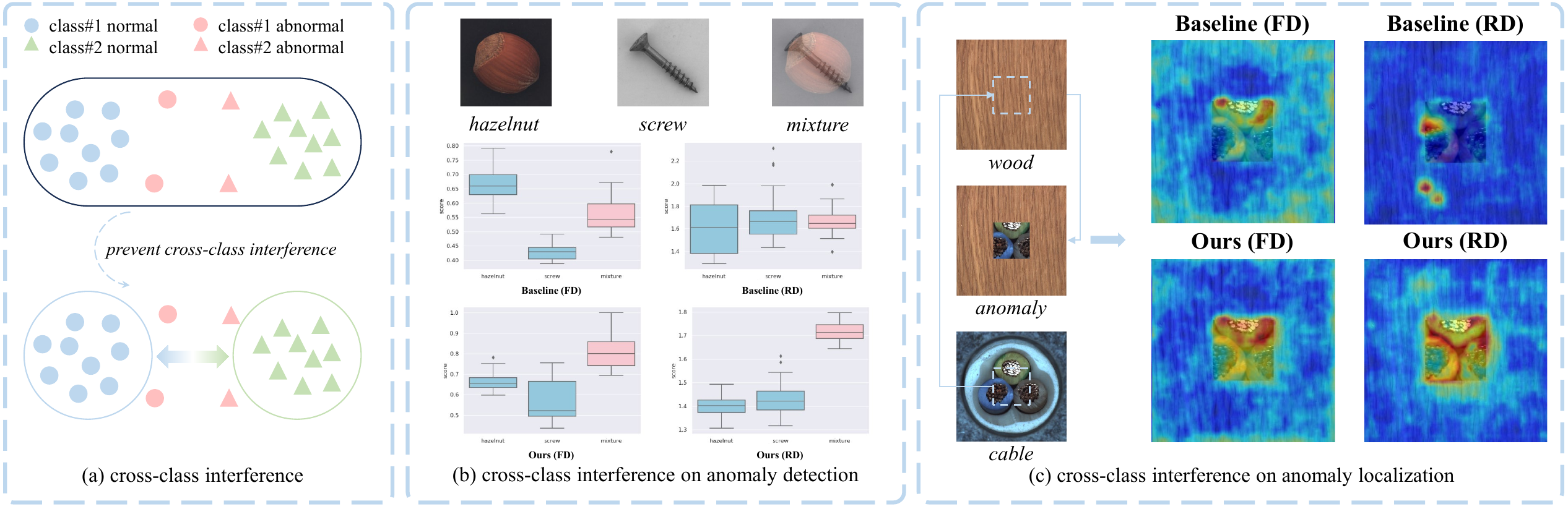}
   \caption{(a) Demonstration of cross-class interference in multi-class anomaly detection. (b) Empirical analysis of cross-class interference. We generated mixtures as anomaly samples from the “hazelnut” and “screw” of MVTecAD via mixup. FD \citep{fd} and RD \citep{rd} show no discrepancy in the anomaly scores, whereas our models exhibit significant differences. (c) Qualitative analysis of cross-class interference. We crop a small region from an image in the “cable” category and paste it onto an image in the “wood” category as an anomaly sample on MVTecAD. Both FD and RD fail to identify synthetic anomalous regions, whereas our models can locate the anomalies precisely.}
   \label{exp}
\end{figure*}

\section{Introduction}

Visual anomaly detection represents a pivotal open-set task in computer vision, aiming to identify unknown anomalous patterns within normal data. This challenge holds significant relevance in a multitude of real-world applications, spanning industrial defect detection \citep{mvtec, visa,jezek2021deep}, video surveillance \citep{video,acsintoae2022ubnormal}, and medical imaging diagnosis \citep{medical, zimmerer2022mood}. 
Traditional anomaly detection approaches often involve training separate models for each specific category. These models are trained on normal samples from their respective categories and can only detect anomalies within the context of that category. While one-class anomaly detection models have shown promise in these contexts \citep{mkd,fd,rd,lee2022cfa}, their inherent limitation lies in the need to construct a separated model for each class, a paradigm that becomes increasingly inefficient with the increasing number of categories. Recent developments have highlighted the emergence of multi-class anomaly detection as a pressing challenge, demanding enhanced scalability and adaptability from anomaly detection models \citep{uniad, zhao2023omnial}. In response to this evolving landscape, we aim to propose a scalable solution for multi-class anomaly detection and localization, where one model can identify anomalies of multiple classes.

Feature reconstruction stands as one of the most influential paradigms in the realm of anomaly detection, distinguished for its robustness and effectiveness. Especially, teacher-student networks become a natural approach for feature reconstruction, involving the prediction of teacher network outcomes through the student network~\citep{kd}. 
In particular, multi-scale distillation is proposed to achieve superior anomaly detection performance by accumulating feature differences between teachers and students under multiple receptive fields \citep{us,mkd,fd}. Recently, by exposing the over-generalization problem on anomaly detection that exists in the forward distillation paradigm, reverse distillation has been proposed as a novel paradigm and achieves SOTA performance on one-class anomaly detection scenarios \citep{rd}. However, we observe substantial performance degradation for both forward distillation \citep{mkd,fd} and reverse distillation \citep{rd} for multi-class anomaly detection, as shown in Fig. \ref{teaser_mvtec}. Therefore, we propose the cross-class interference hypothesis in Fig. \ref{exp}(a), whereby the generalization of the anomaly detection model across different categories causes the model to be somewhat tolerant towards anomalies. 

To empirically assess the impact of cross-class interference on anomaly detection, we conduct two straightforward experiments. In the first experiment, we use Mixup \citep{mixup} technique to superimpose two images belonging to different classes, creating a mixture that should be considered as an anomalous image. We then conduct a statistical analysis of the image-level anomaly scores. As illustrated in Fig. \ref{exp}(b), both forward distillation (FD) \citep{fd} and reverse distillation (RD) \citep{rd} fail to distinguish between mixture and normal images when trained on a multi-class dataset. In the second experiment, we employ a CutPaste-like \citep{li2021cutpaste} anomaly synthesis on images originating from two distinct classes. Accordingly, the synthesized irregularity should be distinguishable for effective anomaly detection models \citep{li2021cutpaste}. However, as shown in Fig. \ref{exp}(c), when training under the multi-class setting, both FD and RD models are unable to identify the anomalous region within the synthesized image.  These experiments demonstrate the detrimental influence of cross-class interference on the performance of teacher-student networks in multi-class anomaly detection and localization.

Evidently, the issue of cross-class interference in multi-class anomaly detection arises from shortcomings in previous teacher-student reconstruction networks, a concern not as prominent in one-class anomaly detection. On the one hand, previous methods primarily train the student network to learn local features from the teacher network without fostering correlations between these features. The absence of such correlations hindered student networks from effectively discerning structural feature differences between the subject and potential anomalies within a sample. Therefore, we propose structural distillation, enabling student networks to discern and capture pairwise feature disparities from teacher networks. In specific, our structural distillation consists of spatial-channel and intra-\&inter-affinity distillation, which represents separate and pairwise feature distances, respectively.
On the other hand, the deficiency of normality constraints leads to weak compactness of multi-class normal representations within teacher-student networks. To tackle this issue, we propose the Central Residual Aggregation Module (CRAM) plugged into the student network. Our proposed CRAM facilitates the learning of compact normality features by aggregating residual projections of student features relative to multiple normality centers. Notably, our multi-class anomaly detection model demonstrates excellent discriminative ability in the experiments presented in Fig. \ref{exp}. 
Overall, we propose Structural Teacher-Student Normality Learning (SNL) to address the problem of \emph{cross-class interference} that hampers the effectiveness of knowledge distillation in multi-class anomaly detection. Notably, our approach offers generalizability to previous teacher-student networks  and improves performance by a large margin in multi-class anomaly detection and localization. Furthermore, our approach remarkably surpasses SOTA on the MVTecAD and VisA datasets. Our main contributions are summarized as follows:

\begin{itemize}

    \item We conduct an in-depth analysis to identify the presence of cross-class interference, which leads to the degradation observed in teacher-student networks when applied to multi-class anomaly detection and localization.
    
    \item To tackle this issue, we propose a structural teacher-student network that learning separate and pairwise feature similarities by spatial-channel and intra-\&inter-affinity distillation.
    
    \item We propose CRAM to be integrated in student network to learn a compact normality representation, thereby enhancing the model's sensitivity to cross-class anomalies.
    
    \item Extensive experiments on the datasets MVTecAD and VisA show that our approach has a dramatic improvement compared to the baseline and also outperforms the state-of-the-art unified models.
    
\end{itemize}

\section{Related Work}
\paragraph{Distillation-based Anomaly Detection:}
Reconstruction is the typical paradigm for anomaly detection, e.g., pixel-level structural reconstruction for industrial defect detection \citep{bergmann2018improving}. Feature-level reconstruction exhibits impressive performance due to the powerful representation capabilities of pre-trained models \citep{yang2020dfr}. Teacher-student networks, which use student networks to reconstruct features of teacher networks, as a natural reconstruction paradigm have been widely studied for anomaly detection. Uninformed student is the first teacher-student network based anomaly detection method \citep{us}. It trains  trains a student network on normal samples to distill from a discriminative teacher network and then detects anomalies by teacher-student differences. The multi-scale knowledge distillation \citep{mkd,fd} is proposed to train a student network to reconstruct the multi-scale features of the teacher network, which is derived from a pre-trained network on ImageNet \cite{deng2009imagenet} with a rich semantic space. In particular, \citep{mkd} utilizes the disparate gradients generated by the model on novel features to detect anomalies and \citep{fd} utilize pyramid reconstruction errors to detect anomalies. In this study, we define this classical teacher-student networks \citep{mkd,fd} as forward distillation. Previous studies have found that forward distillation suffers from anomalous leakage to student networks, whereby more powerful student networks overgeneralize the anomalous representations and thus lead to performance degradation \citep{mkd,fd}. To address this issue, reverse distillation has been proposed to reconstruct shallow multi-scale features progressively from deep features using the student network \citep{rd}, which takes teacher-student networks to state-of-the-art in anomaly detection and localization. Previous approaches have achieved impressive performance on one-class anomaly detection, however, degradation occurs on multi-class anomaly detection. In this study, we aim to achieve high performance in multi-class anomaly detection and localization using teacher-student networks.
\paragraph{Normality Learning:}
DeepSVDD \citep{ruff2018deep} is a one-class normality learning algorithm that detects outliers by training a compact support space for a normality center. Subsequently, the sparse memory mechanism \citep{gong2019memorizing} and the compact memory module \citep{park2020learning} are proposed for learning normality reconstruction. To achieve few-shot adaptation, dynamic normality learning is proposed to project normal prototypes onto a given feature space \citep{lv2021learning}. Recently, CFA \citep{lee2022cfa} proposes the coupled-hypersphere-based feature adaptation to learning normal centers for one-class anomaly detection. In this study, we aim to present a normality learning module that is adaptable to multi-class and sensitive to anomalies.

\begin{figure*}[!ht]
  \centering
   \includegraphics[width=\textwidth]{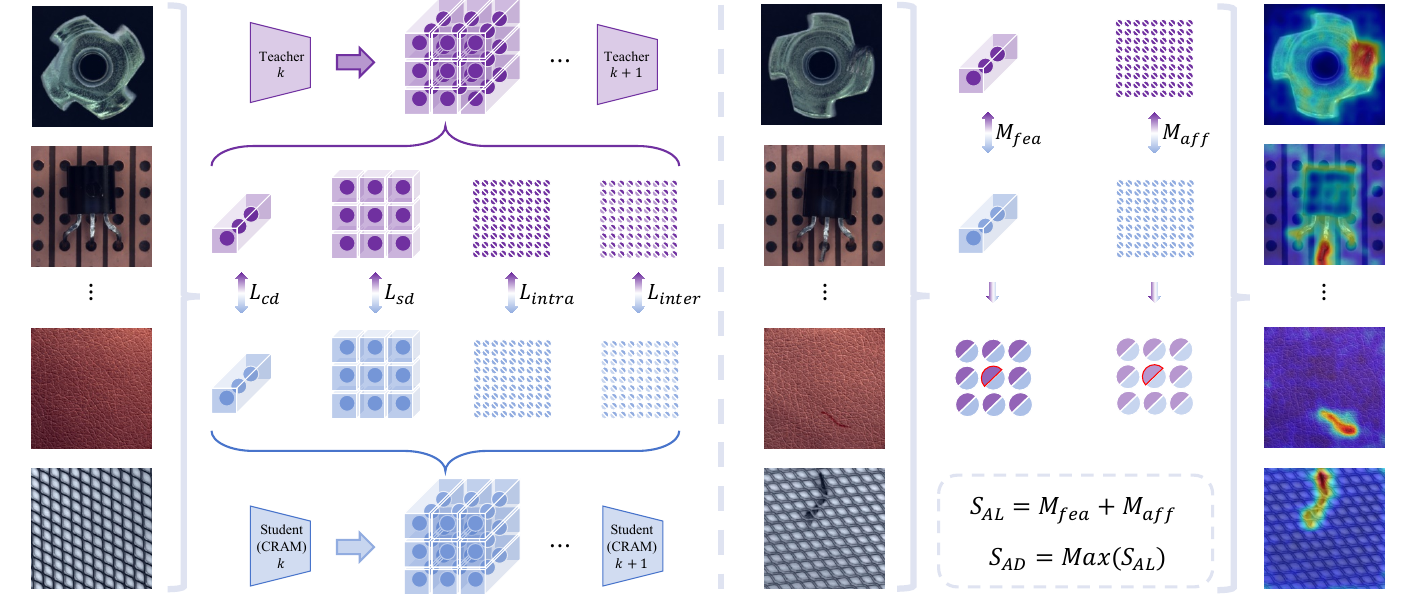}
   \caption{Overview of our structural teacher-student framework. Left: during training with normal samples, our structural distillation quantifies and minimizes the difference between channel-wise features, spatial-wise features, intra-affinity and inter-affinity metrics for the $k$th block of teacher-student network. Right: during testing, for query samples, we measure the local and structural differences respectively by the channel-wise feature distance and intra-affinity distances of the teacher-student network for anomaly detection.}
   \label{framework}
\end{figure*}
\paragraph{Multi-class Anomaly Detection:} UniAD \citep{uniad} initially formulates the task of multi-class visual anomaly detection and proposes a transformer-based feature reconstruction model. Besides, UniAD proposes layer-wise query in the transformer to learn the complex normal distribution of multi-categories. Additionally, OmniAL \citep{zhao2023omnial} proposes a panel-guided approach to synthesize anomalies and train reconstruction and discriminative networks on the synthesized anomaly samples to localize the anomalies. Although the synthetic anomaly approaches \citep{zavrtanik2021draem, zhao2023omnial} provide excellent anomaly localization precision on specific datasets, they require a priori knowledge of the anomalies in the dataset. We commonly define anomalies as unknown so that the model can be sensitive to all kinds of anomalies.

\section{Methodology}
\paragraph{Problem Definition:} For multi-class anomaly detection, we follow a unified setting where the images are from different classes and the category information is inaccessible \cite{uniad}. Let 
$L_{train}=\{I_{normal}^1,...,I_{normal}^n\}$ denotes the set of $n$ anomaly-free training samples from $C$ potential categories. Then, the inference set is defined as 
$L_{test}=\{I_{unknown}^1,...,I_{unknown}^m\}$, which including $m$ query images from the same $C$ classes. Notabaly, the training set $L_{train}$ only includes normal samples and the test set $L_{test}$ includes normal or unknown anomalous samples. 
We aim to achieve a model that can detect anomalous images and localize the anomalous regions in multiple categories.

\paragraph{Preliminaries:}
Lately, the teacher-student networks have made significant strides in advancing anomaly detection \cite{mkd,fd,rd}. In this paradigm, we begin with a pre-trained teacher network capable of extracting rich and discriminative features from images. We train a student network on normal samples to learn and reconstruct these features from the teacher network.  This process is commonly referred to as knowledge distillation \cite{kd}. Subsequently, we use the feature reconstruction errors on query samples to detect anomalies. However, when applied to multi-class anomaly detection, the teacher-student network suffers from degradation, resulting in weak performance. As highlighted earlier, our observations indicate that this performance degradation is attributed to cross-class interference, a phenomenon that affects the model's ability to differentiate anomalies in diverse classes. To overcome this issue, we introduce structural teacher-student normality learning as a novel framework for multi-class anomaly detection and localization. In this section, we present the proposed methodology as follows: (1) structural knowledge distillation, (2) central residual aggregation module for normality learning, and (3) scoring for anomaly detection and localization. These elements collectively form the foundation of our approach, which aims to address the challenge of multi-class anomaly detection by mitigating the impact of cross-class interference.

\subsection{Structural Distillation for Anomaly Detection}
The teacher-student network consists of a frozen pre-trained teacher model and a trainable student model. Particularly, we follow previous work using the same network architecture and distill hierarchical knowledge for the teacher-student network \cite{fd,rd}. Formally, let $F^k_t \in \mathbb{R}^{D^k\times H^k\times W^k}$ and $F^k_s \in \mathbb{R}^{D^k\times H^k\times W^k}$ denote the feature tensors of the $k$th block of the teacher and student models, respectively.  For notation consistency, this paper uses $F_i^k(:,h,w)\in \mathbb{R}^{D^k\times 1}$ to denote the 1-D channel-wise feature at location $(w,h)$ from the feature tensor, and $F_i^k(d,:,:)\in \mathbb{R}^{H^k\times W^k}$ to represent the 2-D spatial feature map in the channel $d$, where $i\in \{t,s\}$. 

During training, the tensor $F^k_t$ extracted from $I_{normal}$ is treated as the learning target. Then, we optimize the student network to produce a reconstructed feature tensor $F^k_s$ that is close to the target tensor $F^k_t$. Following previous works \cite{mkd,fd,kd}, we compute the channel-wise feature distances along the channel axis for the $k$th teacher-student blocks:
\begin{equation}
M^k(h, w) = 1-\frac{(F^k_t(:,h,w))^T\cdot F^k_s(:,h,w)}{\Vert F^k_t(:,h,w)\Vert_2 \cdot \Vert F^k_s(:,h,w)\Vert_2 },
\label{eqn:1}
\end{equation}
where 
$\Vert \cdot \Vert_2$ is the L$2$ norm. By calculate the cosine similarity distance along the channel axis in (1), we obtain a 2-D distance map $M^k\in \mathbb{R}^{H^k\times W^k}$. Considering the hierarchical knowledge distillation, the channel-wise distillation loss is defined as the aggregation of the multi-scale channel-wise distance maps:
\begin{equation}
\mathcal{L}_{cd} = \sum_{k=1}^K[\frac{1}{H^kW^k}\sum_{h=1}^H\sum_{w=1}^W M^k(h, w)],
\label{eqn:cd}
\end{equation}
where $K$ denotes the number of blocks in both teacher and student networks. Note, previous knowledge distillation methods for anomaly detection are typically performed using the channel-wise distance in (\ref{eqn:cd}) \cite{mkd,fd,kd}. 

Apart from encouraging the channel-wise feature consistency, we consider adding spatial feature matching for activation map alignment. Spatial feature distillation refers to having the student network learn the features of the teacher network along a feature map for each dimension. 
We use KL divergence for spatial-wise distillation:
\begin{equation}
\mathcal{L}_{sd} = \sum_k^K \sum_d^D \Phi(F_t^k(d,:,:))\cdot log\frac{\Phi(F_t^k(d,:,:))}{\Phi(F_s^k(d,:,:))},
\label{eqn:sd}
\end{equation}
where $\Phi(\cdot)$ denotes the probability value:
\begin{equation}
\Phi(F^k(d,h,w))=\frac{exp(F^k(d,h,w))}{\sum_h^{H}\sum_w^{ W}exp(F^k(d,h,w))}.
\end{equation}
Unlike the channel-wise loss in (\ref{eqn:cd}) focusing on local consistency, minimizing the KL divergence of the feature maps between teacher-student models in (\ref{eqn:sd}) encourages the global alignment of spatial activations.
By spatial-channel distillation via jointly optimizing $\mathcal{L}_{sd}$ and $\mathcal{L}_{cd}$, we allow the teacher-student network to maintain better local-global continuous consistency. 

It should be noted that in multi-class anomaly detection, the normal distribution across multiple categories becomes significantly more complex than that in one-class scenarios. Due to the cross-class interference, the student model may have more freedom and stronger generalization capabilities in reconstructing abnormal features. While the introduced spatial-wise distillation can somewhat limit the student model's ability to reconstruct globally abnormal features, it does not impose strong constraints on the reconstruction of locally abnormal features, resulting in the failures in Fig. \ref{exp}. To address this issue, we propose structural information distillation. In human vision theory, image structural information describes the inter-dependency between pixels, and these dependencies typically carry crucial information related to objects and semantic understanding \cite{liu2019SKD}. By encouraging alignment of feature tensor's structural information in knowledge distillation, the student model's ability to reconstruct features violating local normality can be significantly reduced. To this end,
let $\mathcal{R}$ denote the reshape function, where $\mathcal{R}(F_i^k)\in \mathbb{R}^{D^k\times H^kW^k}$ for $i\in\{s,t\}$. We use the affinity matrix to represent the structural relation of the feature map:
$
\mathcal{A}^k_s = \mathcal{R}(F_s^k)^T \times \mathcal{R}(F_s^k)
$
and
$
\mathcal{A}^k_t = \mathcal{R}(F_t^k)^T \times \mathcal{R}(F_t^k)
$, and L2 normalization is applied to scale $F_t^k$ and $F_t^k$.
Then, the intra-affinity distillation objective function for the teacher-student network is:
\begin{equation}
\mathcal{L}_{intra} = \sum_k^K \sum_i^{H\cdot W} \sum_j^{H\cdot W} \Vert \mathcal{A}^k_s(i,j) - \mathcal{A}^k_t(i,j) \Vert_2.
\end{equation}
In addition, we impose an external affinity distillation to keep the pairwise similarity consistent across samples. Particularly, the training samples from different categories within a batch assist the student network in learning from the teacher about the discrepancies of representations between classes. 
Our cross affinity matrix is defined as:
$
\mathcal{\tilde{A}}^k_s = \mathcal{R}(F_s^k)^T \times \mathcal{R}(\tilde{F}_s^k)
$
and
$
\mathcal{\tilde{A}}^k_t = \mathcal{R}(F_t^k)^T \times \mathcal{R}(\tilde{F}_t^k).
$
Then the inter-affinity distillation loss is computed as:
\begin{equation}
\mathcal{L}_{inter} = \sum_k^K \sum_i^{H\cdot W} \sum_j^{H\cdot W} \Vert \mathcal{\tilde{A}}^k_s(i,j) - \mathcal{\tilde{A}}^k_t(i,j) \Vert_2.
\end{equation}
Through this, the student network not only distills the pairwise similarity within the sample from the teacher network, but also learns categorical discrepancies.
In summary, the structural distillation objective function for the teacher-student network is:
\begin{equation}
    \mathcal{L} = \mathcal{L}_{cd} + \lambda_1\mathcal{L}_{sd} + \lambda_2\mathcal{L}_{intra} + \lambda_3\mathcal{L}_{inter},
\end{equation}
where the $\lambda_1$, $\lambda_2$, and $\lambda_3$ are the hyper-parameters for the optimization of the teacher-student network.

\subsection{Central Residual Aggregation Module}

\begin{figure}[ht]
  \centering
   \includegraphics[width=0.95\linewidth]{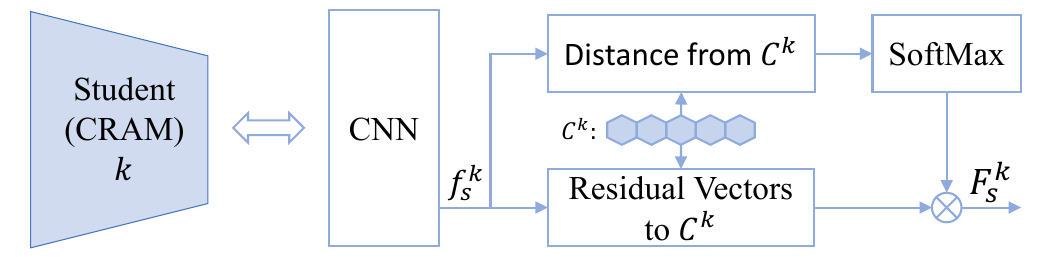}
   \caption{The overview of proposed CRAM. We add CRAM after each CNN block of the baseline student networks \cite{fd,rd}.}
   \label{cram}
\end{figure}

We introduce a Central Residual Aggregation Module (CRAM) to enable the student network to learning the normality pattern as shown in Fig. \ref{cram}. Our CRAM is derived from codebook learning \cite{arandjelovic2016netvlad,yang2022dynamic,wu2021generalized} and includes learnable assignment parameters $\alpha$ and clustering centers $\mathcal{C}\in \mathbb{R}^{D^k\times N}$, where $N$ is the number of clusters. In addition, our CRAM changes the feature aggregation strategy for local feature reorganization. We assume that the normal centers represent finite normal patterns, and that unknown abnormal features are far from the centers. The student model can learn compact normal representations through CRAM and thus produce more significant error responses to abnormal regions with respect to the teacher. Inside a CRAM student block, we denote $f_s^k\in \mathbb{R}^{D^k\times H^k \times W^k}$ as the features output from the CNN block. Since the features are from multiple categorical distributions, we first align the unified centers to the current feature vector $f_s^k(:,h,w)$ through the residual calculation: $r^k(h,w)=f_s^k(:,h,w)-\mathcal{C} $ and $r^k(:,h,w)\in \mathbb{R}^{D^k\times N} $. Then, the soft-assignment formula for aggregating the residual centers is:
\begin{equation}
a_n = \frac{exp(-\alpha\Vert r_n^k(:,h,w) \Vert^2)}{\sum_n^N exp(-\alpha\Vert r_n^k(:,h,w) \Vert^2)},
\end{equation}
where we compute the SoftMax distribution for the distance of features from the centers. We obtain the compact representation of the student network by aggregation:
\begin{equation}
F_s^k(:,h,w) = \sum_n^N a_n r_n^k(:,h,w).
\end{equation}
During the training phase, the learnable centers construct a normal feature space. When anomalies are encountered, the residuals between the features and centers appear different. Thus feature discrepancy in the student-teacher network becomes more significant with CRAM normality learning.

\subsection{Scoring for Anomaly Detection \& localization.}
In the inference phase, we consider reconstruction error and affinity error between student and teacher networks as the measurement. The intuition is that as we minimize the structural distillation loss, the outputs of the teacher and student networks are quite similar for anomaly-free samples. When confronted with unknown features, which are from abnormal samples, our model produces relatively greater losses. First, similar to previous work, we calculate the cosine similarity of the features at each location \cite{mkd,fd,rd}. By Eq. (\ref{eqn:1}), we obtain the distance map $M^k$ from $k$th block of the model. Then, the anomaly map from the separate feature distance $\mathcal{M}_c\in \mathbb{R}^{H\times W}$ is computed as:
\begin{equation}
\mathcal{M}_{fea} = \sum_k^K \mathbb{U}^k(M^k),
\end{equation}
where $\mathbb{U}(\cdot)$ indicates the upsampling operation to resize the anomaly map to the size of the input image $(H,W)$. In addition, to tackle cross-class interference, we utilize the intra-affinity error of the teacher and student features to measure anomalous scores. 
\begin{equation}
\mathcal{E} = \Vert \mathcal{A}^k_s-\mathcal{A}^k_t \Vert_2,
\end{equation}
where $\mathcal{E}\in \mathbb{R}^{H^kW^k\times H^kW^k}$. Then, the pairwise similarity difference map is computed as:
\begin{equation}
\mathcal{M}_{aff} = \mathcal{\tilde{R}} (\sum_k^K \mathbb{U}^k(\frac{1}{H^kW^k}\sum_i^{H^kW^k}\mathcal{E}(:,i) ) ),
\end{equation}
where $\mathcal{\tilde{R}}(\cdot) \in \mathbb{R}^{H \times W} $ denotes the reshape operation. Then, the overall anomaly map is calculated as:
\begin{equation}
S_{AL} = \mathcal{M}_{fea} + \mathcal{M}_{aff},
\end{equation}
where $S_{AL}$ is the pixel-level anomaly map for evaluate the anomaly localization. Additionally, we use the most responsive anomaly score for anomaly detection. Thus the image-level anomaly score is calculated as:
\begin{equation}
S_{AD} = max(S_{AL}),
\end{equation}
where $max(\cdot)$ denotes calculating the maximum value.

\section{Experiment}

\begin{table*}[ht]
\centering
\fontsize{6.5}{15}\selectfont
\renewcommand{\arraystretch}{0.9}
\setlength\tabcolsep{1.5pt}
\begin{tabular}{l|ccccccccccccccc
>{\columncolor[HTML]{DAE8FC}}c }
\hline
MVTecAD  & Bottle   & Cable     & Capsule   & Carpet    & Grid      & Hazelnut  & Leather   & Metal nut & Pill      & Screw     & Tile      & Toothbrush & Transistor & Wood      & Zipper    & Mean                             \\ \hline
UniAD \cite{uniad}    & 99.7/{\color[HTML]{9B9B9B}100} & 95.2/{\color[HTML]{9B9B9B}97.6} & 86.9/{\color[HTML]{9B9B9B}85.3} & 99.8/{\color[HTML]{9B9B9B}99.9} & 98.2/{\color[HTML]{9B9B9B}98.5} & 99.8/{\color[HTML]{9B9B9B}99.9} & 100/{\color[HTML]{9B9B9B}100}   & 99.2/{\color[HTML]{9B9B9B}99.0} & 93.7/{\color[HTML]{9B9B9B}88.3} & 87.5/{\color[HTML]{9B9B9B}91.9} & 99.3/{\color[HTML]{9B9B9B}99.0} & 94.2/{\color[HTML]{9B9B9B}95.0}  & 99.8/{\color[HTML]{9B9B9B}100}   & 98.6/{\color[HTML]{9B9B9B}97.9} & 95.8/{\color[HTML]{9B9B9B}96.7} & 96.5/{\color[HTML]{9B9B9B}96.6} \\
OmniAL \cite{zhao2023omnial}   & 100/{\color[HTML]{9B9B9B}99.4} & 98.2/{\color[HTML]{9B9B9B}97.6} & 95.2/{\color[HTML]{9B9B9B}92.4} & 98.7/{\color[HTML]{9B9B9B}99.6} & 99.9/{\color[HTML]{9B9B9B}100}  & 95.6/{\color[HTML]{9B9B9B}98.0} & 99.0/{\color[HTML]{9B9B9B}97.6} & 99.2/{\color[HTML]{9B9B9B}99.9} & 97.2/{\color[HTML]{9B9B9B}97.7} & 88.0/{\color[HTML]{9B9B9B}81.0} & 99.6/{\color[HTML]{9B9B9B}100}  & 100/{\color[HTML]{9B9B9B}100}    & 93.8/{\color[HTML]{9B9B9B}93.8}  & 93.2/{\color[HTML]{9B9B9B}98.7} & 100/{\color[HTML]{9B9B9B}100}   & 97.2/{\color[HTML]{9B9B9B}97.0}                        \\ \hline
FD \cite{fd}      & 76.8/{\color[HTML]{9B9B9B}100} & 96.5/{\color[HTML]{9B9B9B}95.1} & 78.9/{\color[HTML]{9B9B9B}75.8} & 96.5/{\color[HTML]{9B9B9B}99.4} & 98.8/{\color[HTML]{9B9B9B}99.1} & 99.3/{\color[HTML]{9B9B9B}100}  & 96.2/{\color[HTML]{9B9B9B}97.3} & 97.9/{\color[HTML]{9B9B9B}99.4} & 91.3/{\color[HTML]{9B9B9B}94.1} & 76.8/{\color[HTML]{9B9B9B}93.0} & 99.8/{\color[HTML]{9B9B9B}100}  & 88.9/{\color[HTML]{9B9B9B}99.7}  & 96.2/{\color[HTML]{9B9B9B}96.6}  & 99.4/{\color[HTML]{9B9B9B}99.6} & 79.7/{\color[HTML]{9B9B9B}88.7} & 91.5/{\color[HTML]{9B9B9B}95.9}                        \\
SNL (FD) & 100/{\color[HTML]{9B9B9B}100}  & 98.1/{\color[HTML]{9B9B9B}99.6} & 91.7/{\color[HTML]{9B9B9B}97.4} & 99.9/{\color[HTML]{9B9B9B}100}  & 99.2/{\color[HTML]{9B9B9B}99.9} & 100/{\color[HTML]{9B9B9B}100}   & 100/{\color[HTML]{9B9B9B}100}   & 99.9/{\color[HTML]{9B9B9B}100}  & 95.8/{\color[HTML]{9B9B9B}99.1} & 87.6/{\color[HTML]{9B9B9B}95.3} & 100/{\color[HTML]{9B9B9B}99.6}  & 96.7/{\color[HTML]{9B9B9B}92.5}  & 98.4/{\color[HTML]{9B9B9B}99.6}  & 99.7/{\color[HTML]{9B9B9B}99.4} & 99.1/{\color[HTML]{9B9B9B}97.8} & {\ul 97.7}/{\color[HTML]{9B9B9B}{\ul 98.7}}                            \\
RD \cite{rd}      & 66.5/{\color[HTML]{9B9B9B}100} & 79.3/{\color[HTML]{9B9B9B}95.0} & 93.6/{\color[HTML]{9B9B9B}96.3} & 97.0/{\color[HTML]{9B9B9B}98.9} & 99.0/{\color[HTML]{9B9B9B}100}  & 100/{\color[HTML]{9B9B9B}99.9}  & 100/{\color[HTML]{9B9B9B}100}   & 99.3/{\color[HTML]{9B9B9B}100}  & 95.0/{\color[HTML]{9B9B9B}96.6} & 96.5/{\color[HTML]{9B9B9B}97.0} & 98.7/{\color[HTML]{9B9B9B}99.3} & 99.1/{\color[HTML]{9B9B9B}99.5}  & 92.9/{\color[HTML]{9B9B9B}96.7}  & 99.4/{\color[HTML]{9B9B9B}99.2} & 99.1/{\color[HTML]{9B9B9B}98.5} & 94.4/{\color[HTML]{9B9B9B}98.5}                        \\
SNL (RD) & 100/{\color[HTML]{9B9B9B}100}  & 94.2/{\color[HTML]{9B9B9B}99.1} & 95.4/{\color[HTML]{9B9B9B}97.7} & 98.6/{\color[HTML]{9B9B9B}99.3} & 99.2/{\color[HTML]{9B9B9B}100}  & 100/{\color[HTML]{9B9B9B}100}   & 100/{\color[HTML]{9B9B9B}100}   & 100/{\color[HTML]{9B9B9B}100}   & 95.8/{\color[HTML]{9B9B9B}97.9} & 96.6/{\color[HTML]{9B9B9B}98.1} & 100/{\color[HTML]{9B9B9B}99.7}  & 99.4/{\color[HTML]{9B9B9B}99.4}  & 95.2/{\color[HTML]{9B9B9B}99.6}  & 99.6/{\color[HTML]{9B9B9B}99.2} & 99.7/{\color[HTML]{9B9B9B}98.9} & \textbf{98.3}/{\color[HTML]{9B9B9B}\textbf{99.3}}                        \\ \hline
\end{tabular}
\caption{Unified anomaly detection results with image-level AUROC on MVTecAD. The multi-class/{\color[HTML]{9B9B9B}one-class} performance is reported for each method. The best mean outcome is noted in bold and the runner-up mean outcome is underlined.}
\label{mvtec_sp}
\end{table*}

\begin{table*}[ht]
\centering
\fontsize{7}{15}\selectfont
\renewcommand{\arraystretch}{0.9}
\setlength\tabcolsep{2.78pt}
\begin{tabular}{lcccccccccccc
>{\columncolor[HTML]{DAE8FC}}c }
\hline
VisA    & PCB1      & PCB2      & PCB3      & PCB4      & Macaroni1 & Macaroni2 & Capsules  & Candles   & Cashew    & Chewing gum & Fryum     & Pipe fryum & Mean               \\ \hline
UniAD \cite{uniad}  & 95.4/{\color[HTML]{9B9B9B}95.9} & 93.6/{\color[HTML]{9B9B9B}90.5} & 90.2/{\color[HTML]{9B9B9B}91.0} & 99.4/{\color[HTML]{9B9B9B}98.1} & 93.1/{\color[HTML]{9B9B9B}93.4} & 85.5/{\color[HTML]{9B9B9B}85.3} & 75.3/{\color[HTML]{9B9B9B}79.1} & 96.4/{\color[HTML]{9B9B9B}94.5} & 92.4/{\color[HTML]{9B9B9B}93.6} & 99.4/{\color[HTML]{9B9B9B}98.4}   & 90.8/{\color[HTML]{9B9B9B}89.3} & 97.4/{\color[HTML]{9B9B9B}97.9}  & 92.4/{\color[HTML]{9B9B9B}92.3}          \\
OmniAL \cite{zhao2023omnial}  & 77.7/{\color[HTML]{9B9B9B}96.6} & 81.0/{\color[HTML]{9B9B9B}99.4} & 88.1/{\color[HTML]{9B9B9B}96.9} & 95.3/{\color[HTML]{9B9B9B}97.4} & 92.6/{\color[HTML]{9B9B9B}96.9} & 75.2/{\color[HTML]{9B9B9B}89.9} & 90.6/{\color[HTML]{9B9B9B}87.9} & 86.8/{\color[HTML]{9B9B9B}85.1} & 88.6/{\color[HTML]{9B9B9B}97.1} & 96.4/{\color[HTML]{9B9B9B}94.9}   & 94.6/{\color[HTML]{9B9B9B}97.0} & 86.1/{\color[HTML]{9B9B9B}91.4}  & 87.8/{\color[HTML]{9B9B9B}94.2}          \\ \hline
FD \cite{fd}     & 87.1/{\color[HTML]{9B9B9B}93.8} & 79.0/{\color[HTML]{9B9B9B}89.3} & 79.0/{\color[HTML]{9B9B9B}84.1} & 95.6/{\color[HTML]{9B9B9B}96.7} & 87.0/{\color[HTML]{9B9B9B}93.4} & 69.3/{\color[HTML]{9B9B9B}83.9} & 69.4/{\color[HTML]{9B9B9B}85.2} & 91.6/{\color[HTML]{9B9B9B}96.4} & 92.5/{\color[HTML]{9B9B9B}98.8} & 95.5/{\color[HTML]{9B9B9B}96.8}   & 94.5/{\color[HTML]{9B9B9B}99.5} & 92.6/{\color[HTML]{9B9B9B}99.1}  & 86.1/{\color[HTML]{9B9B9B}93.0}          \\
SNL(FD) & 94.1/{\color[HTML]{9B9B9B}95.0} & 92.3/{\color[HTML]{9B9B9B}93.8} & 90.3/{\color[HTML]{9B9B9B}95.0} & 99.3/{\color[HTML]{9B9B9B}97.4} & 92.2/{\color[HTML]{9B9B9B}93.7} & 73.2/{\color[HTML]{9B9B9B}88.9} & 72.3/{\color[HTML]{9B9B9B}86.9} & 94.9/{\color[HTML]{9B9B9B}96.6} & 93.8/{\color[HTML]{9B9B9B}99.0} & 96.0/{\color[HTML]{9B9B9B}99.3}   & 94.6/{\color[HTML]{9B9B9B}99.5} & 94.9/{\color[HTML]{9B9B9B}99.2}  & 90.7/{\color[HTML]{9B9B9B}95.3}          \\
RD \cite{rd}     & 95.9/{\color[HTML]{9B9B9B}97.1} & 94.4/{\color[HTML]{9B9B9B}97.0} & 92.3/{\color[HTML]{9B9B9B}96.4} & 99.7/{\color[HTML]{9B9B9B}99.8} & 97.8/{\color[HTML]{9B9B9B}97.3} & 85.6/{\color[HTML]{9B9B9B}98.6} & 76.8/{\color[HTML]{9B9B9B}89.5} & 94.2/{\color[HTML]{9B9B9B}94.3} & 92.6/{\color[HTML]{9B9B9B}97.6} & 90.8/{\color[HTML]{9B9B9B}98.4}   & 95.9/{\color[HTML]{9B9B9B}96.2} & 97.2/{\color[HTML]{9B9B9B}94.6}  & {\ul 92.8}/{\color[HTML]{9B9B9B} {\ul 96.4}}    \\
SNL(RD) & 98.1/{\color[HTML]{9B9B9B}97.6} & 94.8/{\color[HTML]{9B9B9B}96.4} & 95.0/{\color[HTML]{9B9B9B}97.3} & 99.9/{\color[HTML]{9B9B9B}99.9} & 96.8/{\color[HTML]{9B9B9B}98.2} & 84.3/{\color[HTML]{9B9B9B}91.9} & 76.1/{\color[HTML]{9B9B9B}91.3} & 94.7/{\color[HTML]{9B9B9B}95.3} & 95.4/{\color[HTML]{9B9B9B}97.8} & 97.6/{\color[HTML]{9B9B9B}98.9}   & 95.9/{\color[HTML]{9B9B9B}96.5} & 99.5/{\color[HTML]{9B9B9B}99.9}  & \textbf{94.0}/{\textbf{\color[HTML]{9B9B9B}96.8}} \\ \hline
\end{tabular}
\caption{Unified anomaly detection results with image-level AUROC on VisA.}
\label{visa_sp}
\end{table*}

\begin{table*}[ht]
\centering
\renewcommand{\arraystretch}{0.9}
\fontsize{6.5}{15}\selectfont
\setlength\tabcolsep{1.5pt}
\begin{tabular}{l|ccccccccccccccc
>{\columncolor[HTML]{DAE8FC}}c }
\hline
MVTec    & Bottle    & Cable     & Capsule   & Carpet    & Grid      & Hazelnut  & Leather   & Metal nut & Pill      & Screw     & Tile      & Toothbrush & Transistor & Wood      & Zipper    & Mean                             \\ \hline
UniAD \citep{uniad}    & 98.1/{\color[HTML]{9B9B9B}98.1} & 97.3/{\color[HTML]{9B9B9B}96.8} & 98.5/{\color[HTML]{9B9B9B}97.9} & 98.5/{\color[HTML]{9B9B9B}98.0} & 98.2/{\color[HTML]{9B9B9B}98.5} & 96.5/{\color[HTML]{9B9B9B}94.6} & 98.8/{\color[HTML]{9B9B9B}98.3} & 94.8/{\color[HTML]{9B9B9B}95.7} & 95.0/{\color[HTML]{9B9B9B}95.1} & 98.3/{\color[HTML]{9B9B9B}97.3} & 91.8/{\color[HTML]{9B9B9B}91.8} & 98.4/{\color[HTML]{9B9B9B}97.8}  & 97.9/{\color[HTML]{9B9B9B}98.7}  & 93.2/{\color[HTML]{9B9B9B}93.4} & 96.8/{\color[HTML]{9B9B9B}96.0} &  96.8/{\color[HTML]{9B9B9B}96.6} \\
OmniAL \citep{zhao2023omnial}   & 99.2/{\color[HTML]{9B9B9B}99.0} & 97.3/{\color[HTML]{9B9B9B}97.1} & 96.9/{\color[HTML]{9B9B9B}92.2} & 99.4/{\color[HTML]{9B9B9B}99.6} & 99.4/{\color[HTML]{9B9B9B}99.6} & 98.4/{\color[HTML]{9B9B9B}98.6} & 99.3/{\color[HTML]{9B9B9B}99.7} & 99.1/{\color[HTML]{9B9B9B}99.1} & 98.9/{\color[HTML]{9B9B9B}98.6} & 98.0/{\color[HTML]{9B9B9B}97.2} & 99.0/{\color[HTML]{9B9B9B}99.4} & 99.4/{\color[HTML]{9B9B9B}99.2}  & 93.3/{\color[HTML]{9B9B9B}91.7}  & 97.4/{\color[HTML]{9B9B9B}96.9} & 99.5/{\color[HTML]{9B9B9B}99.7} &  \textbf{98.3}/{\color[HTML]{9B9B9B}{\ul 97.8}} \\ \hline
FD \citep{fd}       & 97.1/{\color[HTML]{9B9B9B}98.8} & 96.5/{\color[HTML]{9B9B9B}95.8} & 94.8/{\color[HTML]{9B9B9B}98.6} & 98.2/{\color[HTML]{9B9B9B}99.0} & 97.6/{\color[HTML]{9B9B9B}99.0} & 98.6/{\color[HTML]{9B9B9B}98.6} & 98.8/{\color[HTML]{9B9B9B}99.1} & 94.4/{\color[HTML]{9B9B9B}97.2} & 96.8/{\color[HTML]{9B9B9B}97.6} & 87.6/{\color[HTML]{9B9B9B}98.8} & 95.5/{\color[HTML]{9B9B9B}96.9} & 98.1/{\color[HTML]{9B9B9B}99.0}  & 92.4/{\color[HTML]{9B9B9B}81.9}  & 94.0/{\color[HTML]{9B9B9B}96.5} & 96.5/{\color[HTML]{9B9B9B}98.8} & 95.8/{\color[HTML]{9B9B9B}97.0}                        \\
SNL (FD) & 98.2/{\color[HTML]{9B9B9B}99.0} & 97.6/{\color[HTML]{9B9B9B}97.6} & 98.1/{\color[HTML]{9B9B9B}97.1} & 98.3/{\color[HTML]{9B9B9B}99.0} & 97.9/{\color[HTML]{9B9B9B}98.7} & 98.9/{\color[HTML]{9B9B9B}98.8} & 99.0/{\color[HTML]{9B9B9B}99.5} & 96.3/{\color[HTML]{9B9B9B}97.5} & 98.4/{\color[HTML]{9B9B9B}98.7} & 97.8/{\color[HTML]{9B9B9B}98.9} & 94.8/{\color[HTML]{9B9B9B}96.1} & 98.5/{\color[HTML]{9B9B9B}98.5}  & 95.5/{\color[HTML]{9B9B9B}94.2}  & 94.6/{\color[HTML]{9B9B9B}94.7} & 97.3/{\color[HTML]{9B9B9B}98.0} & 97.4/{\color[HTML]{9B9B9B}{\ul 97.8}}                           \\
RD \citep{rd}       & 92.1/{\color[HTML]{9B9B9B}98.7} & 84.7/{\color[HTML]{9B9B9B}97.4} & 98.4/{\color[HTML]{9B9B9B}98.7} & 98.8/{\color[HTML]{9B9B9B}98.9} & 99.1/{\color[HTML]{9B9B9B}99.3} & 99.0/{\color[HTML]{9B9B9B}98.9} & 99.3/{\color[HTML]{9B9B9B}99.4} & 93.5/{\color[HTML]{9B9B9B}97.3} & 98.5/{\color[HTML]{9B9B9B}98.2} & 99.2/{\color[HTML]{9B9B9B}99.6} & 95.9/{\color[HTML]{9B9B9B}95.6} & 98.9/{\color[HTML]{9B9B9B}99.1}  & 87.6/{\color[HTML]{9B9B9B}92.5}  & 96.1/{\color[HTML]{9B9B9B}95.3} & 98.4/{\color[HTML]{9B9B9B}98.2} & 96.0/{\color[HTML]{9B9B9B}{\ul 97.8}}                        \\
SNL(RD) & 98.3/{\color[HTML]{9B9B9B}98.6} & 94.1/{\color[HTML]{9B9B9B}97.8} & 98.7/{\color[HTML]{9B9B9B}98.4} & 98.5/{\color[HTML]{9B9B9B}99.3} & 99.0/{\color[HTML]{9B9B9B}99.1} & 99.2/{\color[HTML]{9B9B9B}99.3} & 99.0/{\color[HTML]{9B9B9B}99.4} & 97.1/{\color[HTML]{9B9B9B}97.8} & 98.7/{\color[HTML]{9B9B9B}98.7} & 99.3/{\color[HTML]{9B9B9B}99.5} & 95.5/{\color[HTML]{9B9B9B}95.9} & 98.9/{\color[HTML]{9B9B9B}99.0}  & 91.6/{\color[HTML]{9B9B9B}94.0}  & 95.7/{\color[HTML]{9B9B9B}95.6} & 98.5/{\color[HTML]{9B9B9B}{\ul 97.8}} & {\ul 97.5}/{\color[HTML]{9B9B9B} \textbf{98.0}}                        \\ \hline
\end{tabular}
\caption{Unified anomaly localization results with pixel-level AUROC on MVTecAD.}
\label{mvtec_px}
\end{table*}

\begin{table*}[ht]
\centering
\fontsize{7}{15}\selectfont
\renewcommand{\arraystretch}{0.9}
\setlength\tabcolsep{2.78pt}
\begin{tabular}{lcccccccccccc
>{\columncolor[HTML]{DAE8FC}}c }
\hline
VisA    & PCB1      & PCB2      & PCB3      & PCB4      & Macaroni1 & Macaroni2 & Capsules  & Candles   & Cashew    & Chewing gum & Fryum     & Pipe fryum & Mean      \\ \hline
UniAD \cite{uniad}  & 99.3/{\color[HTML]{9B9B9B}99.2} & 97.9/{\color[HTML]{9B9B9B}96.7} & 98.4/{\color[HTML]{9B9B9B}98.0} & 97.9/{\color[HTML]{9B9B9B}98.8} & 99.3/{\color[HTML]{9B9B9B}98.9} & 98.0/{\color[HTML]{9B9B9B}97.1} & 98.3/{\color[HTML]{9B9B9B}98.6} & 99.1/{\color[HTML]{9B9B9B}98.9} & 98.5/{\color[HTML]{9B9B9B}99.2} & 99.1/{\color[HTML]{9B9B9B}98.5}   & 97.6/{\color[HTML]{9B9B9B}97.8} & 99.1/{\color[HTML]{9B9B9B}99.4}  & {\ul 98.5}/{\color[HTML]{9B9B9B}{\ul 98.4}} \\
OmniAL \cite{zhao2023omnial} & 97.6/{\color[HTML]{9B9B9B}98.7} & 93.9/{\color[HTML]{9B9B9B}83.2} & 94.7/{\color[HTML]{9B9B9B}98.4} & 97.1/{\color[HTML]{9B9B9B}98.5} & 98.6/{\color[HTML]{9B9B9B}98.9} & 97.9/{\color[HTML]{9B9B9B}99.1} & 99.4/{\color[HTML]{9B9B9B}98.6} & 95.8/{\color[HTML]{9B9B9B}90.5} & 95.0/{\color[HTML]{9B9B9B}98.9} & 99.0/{\color[HTML]{9B9B9B}98.7}   & 92.1/{\color[HTML]{9B9B9B}89.3} & 98.2/{\color[HTML]{9B9B9B}99.1}  & 96.6/{\color[HTML]{9B9B9B}96.0} \\ \hline
FD \cite{fd}     & 99.2/{\color[HTML]{9B9B9B}99.7} & 97.3/{\color[HTML]{9B9B9B}97.7} & 97.8/{\color[HTML]{9B9B9B}98.1} & 98.3/{\color[HTML]{9B9B9B}97.8} & 97.9/{\color[HTML]{9B9B9B}98.7} & 98.1/{\color[HTML]{9B9B9B}98.7} & 96.7/{\color[HTML]{9B9B9B}97.9} & 98.9/{\color[HTML]{9B9B9B}97.1} & 76.0/{\color[HTML]{9B9B9B}99.0} & 97.9/{\color[HTML]{9B9B9B}98.3}   & 96.6/{\color[HTML]{9B9B9B}95.6} & 98.8/{\color[HTML]{9B9B9B}98.6}  & 96.1/{\color[HTML]{9B9B9B}98.1} \\
SNL(FD) & 99.6/{\color[HTML]{9B9B9B}99.7} & 98.6/{\color[HTML]{9B9B9B}98.3} & 98.5/{\color[HTML]{9B9B9B}98.6} & 98.5/{\color[HTML]{9B9B9B}98.8} & 98.7/{\color[HTML]{9B9B9B}99.0} & 98.3/{\color[HTML]{9B9B9B}99.0} & 97.9/{\color[HTML]{9B9B9B}97.9} & 99.1/{\color[HTML]{9B9B9B}97.8} & 98.4/{\color[HTML]{9B9B9B}99.1} & 98.0/{\color[HTML]{9B9B9B}98.5}   & 97.7/{\color[HTML]{9B9B9B}92.5} & 99.0/{\color[HTML]{9B9B9B}99.1}  & {\ul 98.5}/{\color[HTML]{9B9B9B}98.2} \\
RD \cite{rd}     & 99.5/{\color[HTML]{9B9B9B}99.7} & 97.8/{\color[HTML]{9B9B9B}98.0} & 98.6/{\color[HTML]{9B9B9B}99.3} & 98.2/{\color[HTML]{9B9B9B}98.3} & 99.6/{\color[HTML]{9B9B9B}99.6} & 99.1/{\color[HTML]{9B9B9B}99.4} & 98.0/{\color[HTML]{9B9B9B}99.6} & 98.7/{\color[HTML]{9B9B9B}98.5} & 75.2/{\color[HTML]{9B9B9B}93.5} & 93.9/{\color[HTML]{9B9B9B}98.2}   & 97.5/{\color[HTML]{9B9B9B}97.1} & 99.2/{\color[HTML]{9B9B9B}99.3}  & 96.2/{\color[HTML]{9B9B9B}{\ul 98.4}} \\
SNL(RD) & 99.7/{\color[HTML]{9B9B9B}99.7} & 98.0/{\color[HTML]{9B9B9B}98.3} & 98.2/{\color[HTML]{9B9B9B}99.1} & 98.3/{\color[HTML]{9B9B9B}98.7} & 99.6/{\color[HTML]{9B9B9B}99.6} & 99.2/{\color[HTML]{9B9B9B}99.4} & 98.0/{\color[HTML]{9B9B9B}99.4} & 99.2/{\color[HTML]{9B9B9B}98.7} & 98.4/{\color[HTML]{9B9B9B}95.8} & 98.4/{\color[HTML]{9B9B9B}97.7}   & 97.5/{\color[HTML]{9B9B9B}96.8} & 99.3/{\color[HTML]{9B9B9B}99.3}  & \textbf{98.7}/{\color[HTML]{9B9B9B}\textbf{98.5}} \\ \hline
\end{tabular}
\caption{Unified anomaly localization results with pixel-level AUROC on VisA.}
\label{visa_px}
\end{table*}

\subsection{Dataset}
We evaluate the proposed method on two large-scale visual anomaly detection datasets: MVTecAD \cite{mvtec} and VisA \cite{visa}. MVTecAD is a comprehensive anomaly detection benchmark that includes 10 object categories and 5 texture categories. There are total 3629 normal images for training and 1725 unknown images for testing. VisA contains object images with more complex structures and multiple instances. It includes 9621 normal images and 1200 anomalous images from 12 categories. Note that we train a unified model on data from all categories and evaluate the anomaly detection performance across categories.
\subsection{Implementation Details}
All images in our experiments are resized to 256x256 and normalized by the mean and variance of ImageNet \cite{deng2009imagenet}. We default to using WideResNet-50 \cite{zagoruyko2016wide} as the backbone, where the teacher network loads the model pre-trained on ImageNet. For forward distillation, our framework is based on STPM \cite{fd}, which uses the same teacher and student architectures. For reverse distillation, we follow the structure used in RD \cite{rd}. In this study, we apply the same processing and training strategies to FD-based and RD-based methods. During training, we use the Adam optimizer with a learning rate of $0.005$ and a batch size of 8. For hyper-parameters, we simply set $\lambda_1=\lambda_2=\lambda_3=1$, and the cluster number $N=50$. The experiments are implemented using Pytorch \cite{paszke2019pytorch} on a single RTX3090 GPU. We use image-level and pixel-level Area Under the Receiver Operator Curve (AUROC) to evaluate the anomaly detection and localization performance following UniAD \cite{uniad}.

\begin{table}[!ht]
\centering
\fontsize{9}{11}\selectfont
\setlength\tabcolsep{2pt}
\begin{tabular*}{0.9\linewidth}{cccc|c|
>{\columncolor[HTML]{DAE8FC}}c
>{\columncolor[HTML]{DAE8FC}}c}
\hline
\multicolumn{4}{c|}{Learning Objective}               & Module & \multicolumn{2}{c}{\cellcolor[HTML]{DAE8FC}Evaluation} \\ \hline
$L_{cd}$ & $L_{sd}$ & $L_{intra}$ & $L_{inter}$ & CRAM   & image-level                & pixel-level               \\ \hline
\checkmark            & -            & -              & -              & -      & 94.4                       & 96.0                      \\
\checkmark            & \checkmark            & -              & -              & -      & 94.7                       & 96.4                      \\
\checkmark            & \checkmark            & \checkmark              & -              & -      & 97.4                       & 97.0                      \\
\checkmark            & \checkmark            & \checkmark              & \checkmark              & -      & 97.7                       & 97.1                      \\
\checkmark            & \checkmark            & -              & -              & \checkmark      & 97.1                       & 97.3                      \\
\checkmark            & \checkmark            & \checkmark              & \checkmark              & \checkmark      & 98.3                       & 97.5                      \\ \hline
\end{tabular*}
\caption{Ablation studies of proposed methods. The evaluation is based on RD \cite{rd} and all results are evaluated on MVTecAD.}
\label{ablation1}
\end{table}

\begin{figure}[!ht]
  \centering
   \includegraphics[width=\linewidth]{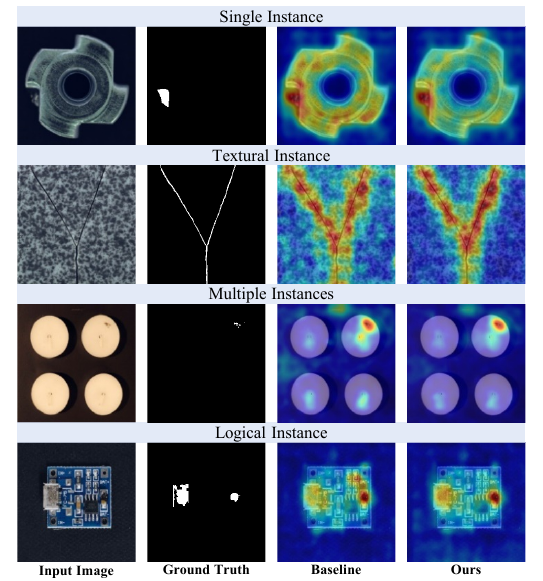}
   \caption{Visualization of our approach and baseline RD \cite{rd} in various anomaly scenarios of MVTecAD and VisA.}
   \label{vis}
\end{figure}

\subsection{Main Results}
The unified models, UniAD \cite{uniad} and OmniAL \cite{zhao2023omnial}, which are specifically proposed for multi-class anomaly detection, are the primary comparators of our approach. In addition, we reproduce the performance of forward distillation (FD) \cite{fd} and reverse distillation (RD) \cite{rd} on multi-class anomaly detection as the baselines. We include a detailed table comparing the performance of other models on multi-class anomaly detection in the supplement \cite{us,yi2020patch,defard2021padim,li2021cutpaste,mkd,zavrtanik2021draem}. We evaluate anomaly detection with image-level AUROC in Tables \ref{mvtec_sp}\&\ref{visa_sp} and anomaly localization with pixel-level AUROC in Tables \ref{mvtec_px}\&\ref{visa_px} for the MVTecAD and VisA datasets. Notably, we not only report the results of multi-class anomaly detection, but also evaluate on one-class anomaly detection as a reference.

\paragraph{Anomaly Detection.}
We show the anomaly detection results on MVTecAD in Tab. \ref{mvtec_sp}. It is obvious to notice the degradation of FD and RD in multi-class anomaly detection. Meanwhile, our proposed SNL greatly improves the performance of FD and RD by 6.2\% and 3.9\%, respectively. Similarly, as shown in Tab. \ref{visa_sp} performance degradation is observed on the VisA dataset, while SNL achieves 4.6\% and 1.2\% improvement compared to FD and RD. Remarkably, we outperform the reconstruction-based SOTA method UniAD by 1.8\% on MVTecAD and 1.6\% on VisA. For OmniAL, a unified model based on anomaly synthesis, our method slightly gains 1.1\% on MVTecAD while exceeds a large margin by 6.2\% on VisA. For one-class tasks, our SNL achieves the best two results based on FD and RD, which shows the generalizability of our method. 

\paragraph{Anomaly Localization.} As shown in Tabs. \ref{mvtec_px}\&\ref{visa_px}, our SNL significantly boosts the anomaly localization capability of FD and RD on MVTecAD and VisA. Compared to UniAD, our SNL(RD) obtains an increase of 0.7\% and 0.2\% on MVTecAD and VisA, respectively. Although the localization performance of our method is slightly lower than OmniAL on MVTecAD, we dramatically outperform OmniAL by 2.1\% on VisA. Anomaly synthesis lacks the generalization ability because it requires certain a prior knowledge. Therefore, a potential future study is to fuse anomaly synthesis into teacher-student networks to enhance localization ability such as in \cite{tien2023revisiting}. Further more, our SNL(RD) achieve the best one-class anomaly localization performance. 

\paragraph{Visualization.} We show in Fig. \ref{vis} a quantitative comparison of the localization results of our method with RD \cite{rd} on various anomaly scenarios in multi-class anomaly detection. Cross-class interference does not mean that the model loses the ability to localize anomalies, but rather lacks sensitivity to anomalous regions. Thus we can observe that the baseline model RD is able to coarsely localize anomalies, but not as accurately as our approach. Furthermore, such imprecision directly leads to the lack of discriminative image-level anomaly scores, which is only slightly reflected in the pixel-level anomaly scores. More visualization results are in the supplement.

\begin{table}[ht]
\centering
\begin{tabular}{l|llll}
\hline
Capacity    & w/o  & 25   & 50            & 75   \\ \hline
image-level & 97.7 & 98.2 & \textbf{98.3} & 98.0 \\
pixel-level & 97.1 & 97.5 & \textbf{97.5} & 97.4 \\ \hline
\end{tabular}
\caption{Ablation study on the number of CRAM centers.}
\label{ablation2}
\end{table}

\subsection{Ablation Study}
We conduct detailed ablation experiments to specifically evaluate the results of our approaches. As shown in Tab. \ref{ablation1}, we analyze the effects of the structural distillation objective functions and the CRAM normality representation. First, we decompose the feature reconstruction into channel-wise and spatial-wise learning to show the gains of spatial feature distillation. Then, intra-\&inter-affinity distillation, which facilitates the teacher-student network to capture pairwise feature similarities, brings 3.0\% and 0.7\% improvement in image-level and pixel-level anomaly recognition performance. In addition, our normality learning module, CRAM, enables the student network to learn compact normal representations, resulting in increased outcomes of 2.4\% and 0.9\% compared to baseline. Overall, our SNL boosts the baseline by 3.9\%/1.5\% of anomaly detection/localization performance. 

For CRAM, we assess the effect of the number of normality centers on the results in Tab. \ref{ablation2}. We observe a significant improvement with the CRAM, but a larger codebook capacity doesn't yield further improvement. Ablation studies regarding model capacities, number of network blocks, etc. can be found in the supplement.

\section{Conclusion}
In this study, we identify \emph{cross-class interference} problem in the teacher-student network for multi-class anomaly detection, which severely reduces the discriminitive ability to anomalies. To overcome this obstacle, we propose structural teacher-student normality learning, which consists of structural distillation and the central residual aggregation module (CRAM). Specifically, structural distillation presents feature affinity reconstruction for capturing pairwise feature similarities between teacher and student features. Besides, CRAM assists student networks in learning discriminative normality representations. Extensive experiments show that our method dramatically improves baseline results and outperforms state-of-the-art methods.

{
    \small
    \bibliographystyle{ieeenat_fullname}
    \bibliography{main}
}


\end{document}